# Regional Priority based Anomaly Detection using Autoencoders

**Shruti Mittal, Dattaraj Rao**
General Electric

## ABSTRACT

In the recent times, autoencoders, besides being used for compression, have been proven quite useful even for regenerating similar images or help in image denoising. They have also been explored for anomaly detection in a few cases. However, due to location invariance property of convolutional neural network, autoencoders tend to learn from or search for learned features in the complete image. This creates issues when all the items in the image are not equally important and their location matters. For such cases, a semi supervised solution - regional priority based autoencoder (RPAE) has been proposed. In this model, similar to object detection models, a region proposal network identifies the relevant areas in the images as belonging to one of the predefined categories and then those bounding boxes are fed into appropriate decoder based on the category they belong to. Finally, the error scores from all the decoders are combined based on their importance to provide total reconstruction error.

## I. INTRODUCTION

Locomotives use a camera to record track videos to help them investigate in case of an accident. These track videos can also be used to identify visible defects. But the frequency of seeing such defects is pretty low and it becomes hard to train a machine learning model for them. Our previous paper Ref [2] covers attempts on using supervised learning to identify some of these track defects. In order to handle defects for which supervised classifier with satisfactory accuracy cannot be obtained, we must depend on traditional image processing methods which do not generalize very well and take a lot of time to build safeguards around each and individual specific case. To solve this issue, autoencoders were explored. They utilize unsupervised learning and learn to reproduce the images that they are trained on. Since track videos largely contains good healthy track, model would learn to reproduce them better and the error for a generated healthy track would be much less than that for an anomalous one, helping to raise an alert for checking the issue.

## II. CONVENTIONAL AUTOENCODER

The method first attempted for track anomaly detection - autoencoders refer to a kind of neural networks architecture and are typically used for efficient compression. Neural network models have been traditionally used for classification or regression. As they consist of many nonlinear units arranged in multiple layers, they have been found to be able to learn very complicated features easily. Since neural networks were so good at learning features, autoencoders were proposed to learn a good compressed representation.

**Model Description**
These models have two parts – encoders and usually its mirror image as decoders. They are trained with outputs same as the inputs. Its architecture forces an autoencoder to first produce an efficient, compressed representation and then try to recreate the input from that representation. For the compression, once trained, only encoder part is used and the decoder part is used separately to retrieve complete image when required.

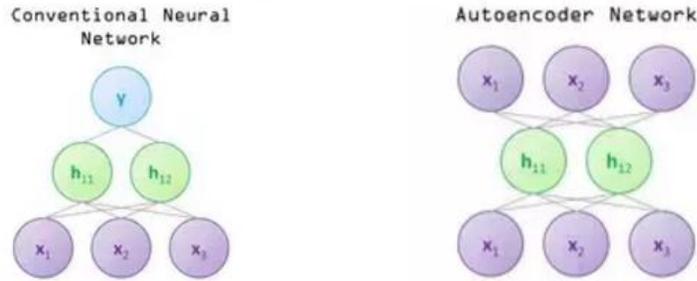

Figure 1: Simple neural network vs Autoencoders (Image Source: https://www.quora.com/What-is-the-difference-between-a-neural-network-and-an-autoencoder-network)

Now if this model has been trained to reproduce a certain kind of inputs, while training, its weights will get optimized to reproduce those type of features and won't work that effectively if a completely different input is provided. In the image context, if the model has been trained to recreate images of digit 1, the weights will get optimized to reproduce 1 pretty well. However, if given a very noisy image of 1, or if given an image of 2, the model would be inclined towards ignoring new features and won't be including them in the recreated image. This results in autoencoders being used also for denoising the original images.

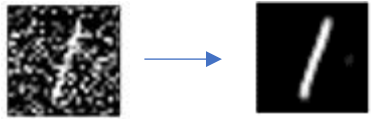

*Figure 2*: Recreated Image from an autoencoder trained to reproduce 1s (Image Source: https://blog.keras.io/building-autoencoders-in-keras.html)

**Auto Encoders for Railway Track**

A similar approach was attempted to detect railway track anomalies. As almost all the track videos contain healthy track images, autoencoder should learn to reproduce them well and if there is any defect like sunkink, vegetation overgrowth, broken track, missing tie or some obstruction, model won't be able to reproduce it that well and the output image would look different from the input one. If this difference is more than a threshold, it can be flagged as an anomaly. Based on the percentage of difference between input image and recreated output image, a track health index can be assigned which will provide an idea about the severity of the anomaly.

Initial attempts showed models can easily recreate the images, a few samples of which are shown in Figure 3. The issue was deciding which one is an anomaly. Reconstruction error easily gets impacted by different surrounding and ambience conditions. The presence of a sunkink on track gets neglected because of occurrence of a tunnel or small hill by the side.

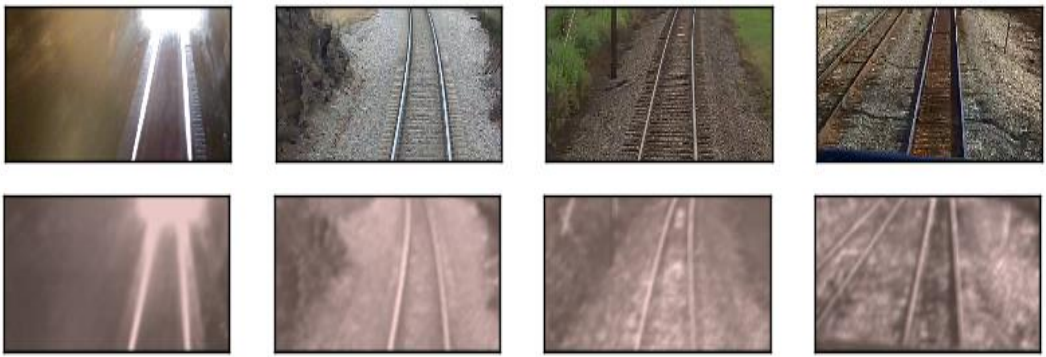

*Figure 3*: Reconstructed Track Images

**Failure Scenarios**
Based on these initial attempts, two major failure scenarios were identified for conventional autoencoders.

1. They give equal importance to a feature identified in any part of the image. An example of such a case is broken track. If the track on which loco is moving is broken, that is an anomaly, however if a broken track segment is lying by the side of the main track, that needs to be ignored. Similarly, if vegetation is there by the side of the track, that is alright, however if it is in between the railway tracks, it interferes with the uniform load distribution and should be flagged as an anomaly.

2. They consider all the anomalous features at par but to flag track anomalies, not all types of anomalies are equally important. If we simply consider a single threshold of reconstruction error for all the different anomalies, a lot of vegetation all through the track will make our image a lot more different than just a small broken piece of track. Different track anomalies have very different priorities and an issue with track alignment needs to have much more weightage than a bad ballast or a tie issue.

To resolve both these issues, a model is needed which can identify the important regions in the image, prioritize them and then focus on reconstructing only selected image segments instead of complete image. This brings into picture region and priority based anomaly detection as explained in the next section.

### III. PROPOSED SOLUTION

The first task to resolve is to be able to identify region of interests (ROI). Similar to object detection models, an embedded region proposal network to recommend subsections to be regenerated is used for that. As first proposed in Ref [1], an RPN, using the feature maps extracted by a CNN based model, predicts bounding box coordinates along with a probability score that segment contained is the region of interest. For object classification, these coordinates and probability score is then passed on to the classifier, which also takes in relevant section of the initially extracted feature map. As input to the classifier needs to have a fixed size and the object bounding box can vary in size, ROI pooling is used, which basically does max pooling with feature map for ROI as the input and the predefined size of classifier input as its output. Architecture of such a model is shown in Figure 4.

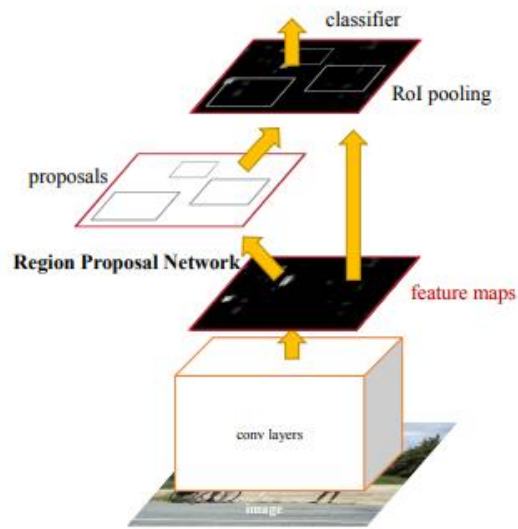

*Figure 4*: Faster-RCNN Object Detection Model Architecture [Ref 1]

The overall architecture of model proposed is shown in Figure 5. It will first extract features to build an encoded representation. This encoded layer will act as input feature for region proposal network (RPN). RPN will propose bounding boxes based on 3 anchor boxes. For all the regions proposed, corresponding decoder will be used to reproduce that specific subsection of the original image.

Reconstruction error will be calculated only for reproduced subsection. All the reconstruction errors will then be added over, weighted by the priority assigned to their anchor box, based on defect seriousness, to calculate a Track Health Index (THI) value. THI will be the final judge to decide when to flag an anomaly. While training, cost function will also depend on THI. And since regions focused on track segments have higher weightage in error calculation, model will pay more importance to learning exact reproduction of those segments as compared to other areas.

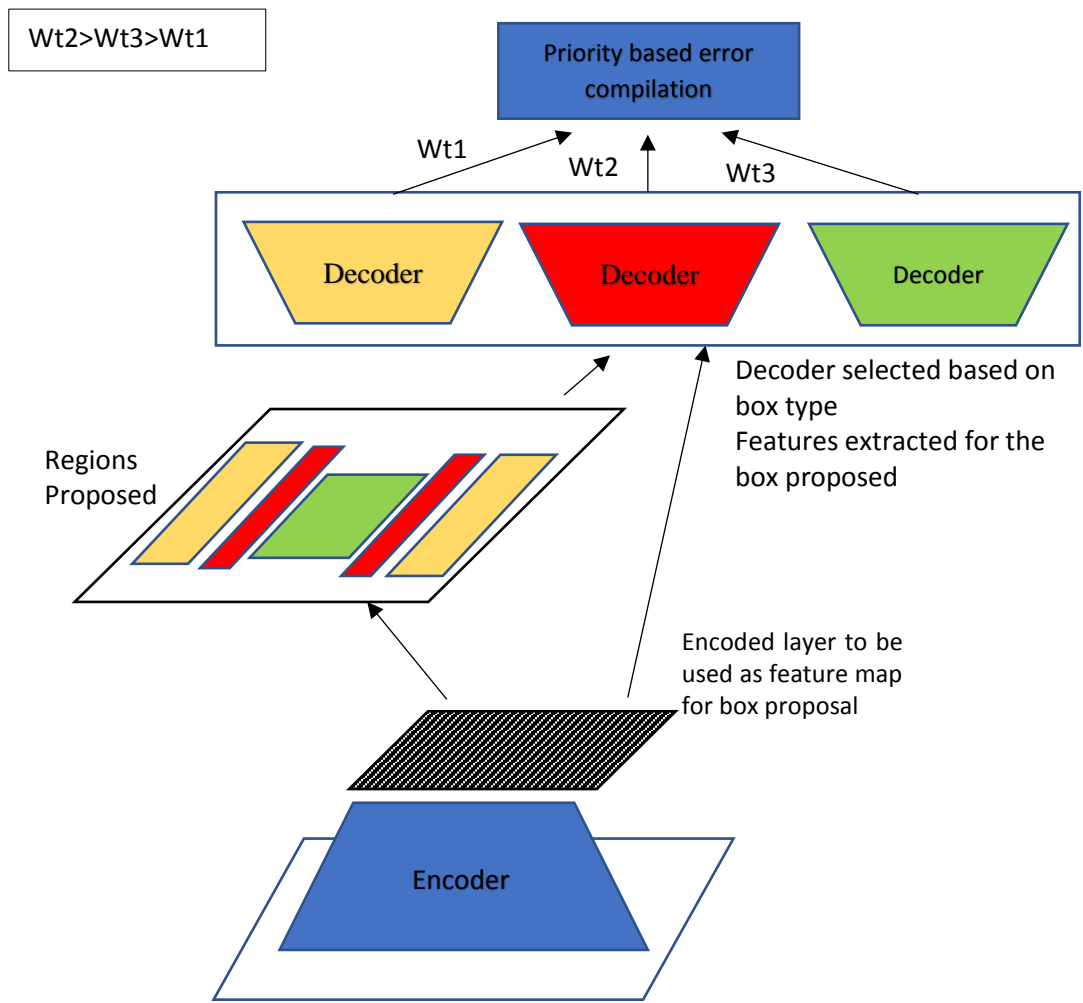

*Figure 5:* Regional Priority based Autoencoder - Model Architecture

## IV. CONCLUSION

In this paper we have proposed a semi supervised option regional priority based autoencoder (RPAE) for anomaly detection. It provides a solution for cases where sufficient training data is not available for training a classifier, but a completely unsupervised approach for anomaly detection is also not possible as location of anomalous feature detected matters and also different anomalies have different weightage. It utilizes the RPNs used for object detection in Faster RCNN to identify relevant regions and then uses appropriate decoder. The use case shown here is identification of railway track issues.